\documentclass{article}
\pdfoutput=1
\usepackage{subcaption}
\usepackage{amssymb}
\usepackage{amsmath}
\usepackage{booktabs}
\usepackage{multirow}
\usepackage{graphicx}
\usepackage{float} 
\usepackage{wrapfig}
\usepackage[linesnumbered, ruled, vlined]{algorithm2e}
\usepackage{placeins}
\usepackage{enumitem}
\usepackage{url}

\usepackage[final]{corl_2025} 

\usepackage{color}

\title{Multimodal Fused Learning for Solving the Generalized Traveling Salesman Problem in Robotic Task Planning}

\author{
\begin{tabular}{c}
Jiaqi Cheng$^{1}$ \quad 
Mingfeng Fan$^{2}$\thanks{Corresponding author: ming.fan@nus.edu.sg} \quad 
Xuefeng Zhang$^{2}$ \quad 
Jingsong Liang$^{2}$ \quad 
Yuhong Cao$^{2}$ \\
Guohua Wu$^{1}$ \quad 
Guillaume Adrien Sartoretti$^{2}$\\
\textnormal{$^{1}$Central South University}\quad 
\textnormal{$^{2}$National University of Singapore} 
\end{tabular}
}

\begin{document}
\maketitle


\begin{abstract}
    Effective and efficient task planning is essential for mobile robots, especially in applications like warehouse retrieval and environmental monitoring. These tasks often involve selecting one location from each of several target clusters, forming a Generalized Traveling Salesman Problem (GTSP) that remains challenging to solve both accurately and efficiently. To address this, we propose a Multimodal Fused Learning (MMFL) framework that combines graph-based topology with image-based spatial representations to develop effective real-time task planning solutions. Specifically, we first introduce a novel coordinate-to-image builder that converts GTSP problem instances into spatially informative representations, complemented by an adaptive resolution scaling mechanism that ensures consistent performance across varying problem scales. We then incorporates a multimodal fusion mechanism featuring dedicated bottleneck that effectively merge topological and geometric information streams. Extensive experiments show that our MMFL approach significantly outperforms state-of-the-art methods across various GTSP instances while maintaining the computational efficiency required for real-time robotic applications. Physical robot tests further validate its practical effectiveness in real-world scenarios\footnote{The source code is available at \url{https://github.com/Carveller/MMFL-for-GTSP}}.
    
\end{abstract}



\section{Introduction} \label{sec:intro}

\begin{figure*}[htbp]
    \vspace{-0.8cm}
    \centering 
    \setlength{\abovecaptionskip}{0.cm}
    \includegraphics[width=\textwidth]{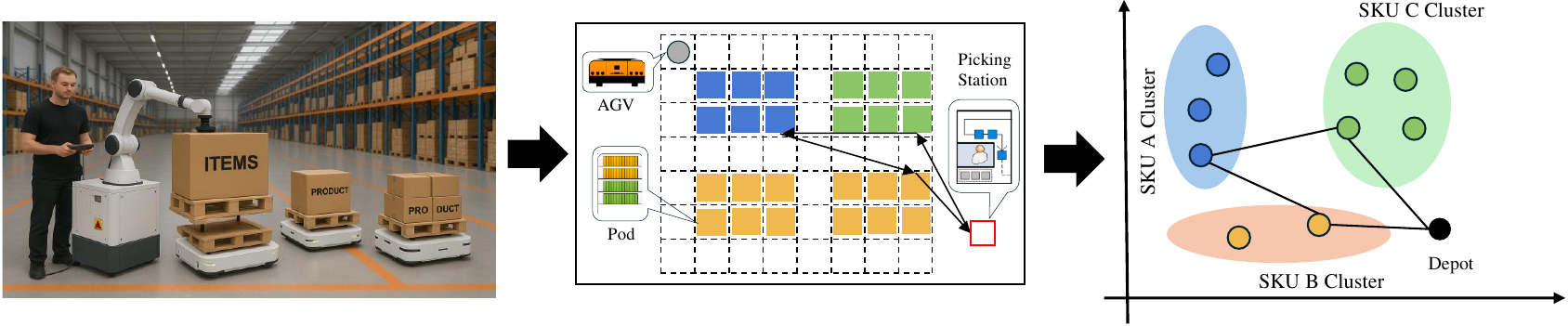} 
    \caption{Representation of GTSP in Warehouse Environment.} 
    \label{representation}
    \vspace{-0.5cm}
\end{figure*}

Autonomous mobile robots have become indispensable in warehouse logistics~\cite{qiu2025integrated, zhang2025real}, healthcare, manufacturing~\cite{sandrini2025learning}, and disaster-response operations~\cite{huang2024hierarchical}, where they are expected to complete complex, multi-stop missions while minimizing travel distance, energy expenditure, and makespan~\cite{bai2025order, zhang2025bi, kim2025optimized, yu2025deep}. A common requirement in these settings is that a robot must service exactly one location from each of several logically defined item or inspection sets. In a warehouse, for instance, a single stock-keeping unit (SKU) may reside on multiple shelves distributed throughout the facility. The robot may choose any one of those shelves as long as every requested SKU is collected~\cite{adachi2024path, zahoran2022proseqqo, bock2025survey}. This requirement is naturally formulated as a Generalized Traveling Salesman Problem (GTSP), in which nodes are partitioned into clusters that represent interchangeable task alternatives and the objective is to find the shortest tour visiting one node from every cluster~\cite{pop2024comprehensive}, as shown in Figure~\ref{representation}.

While exact branch-and-bound or cutting-plane algorithms can guarantee optimal solutions to GTSP, their computational cost scales exponentially with the problem size, making them impractical for on-board execution in real-time robotic applications \cite{deckerova2024combinatorial, fischetti1997branch, yuan2020branch}. Metaheuristic techniques such as genetic algorithms~\cite{ahmed2024effects}, ant-colony optimization~\cite{kato2025method}, and memetic algorithm (MA)~\cite{cosma2024novel} offer better scalability but require extensive parameter tuning and domain-specific adaptation, which severely limits their portability across different robots and environments. More recently, neural approaches have shown promise by learning heuristics directly from data \cite{chung2025neural}. These neural-based approaches can be categorized into graph-learning methods and hybrid learning–search methods. In neural-based approaches, most of these models only take in graph-based inputs. For example, POMO~\cite{kwon2020pomo}, which employs a Transformer-style graph encoder as our model, outperforms the edge-aware GCN~\cite{joshi2019efficient}. In addition, several other methods~\cite{fang2024learning, kong2024efficient, kool2018attention, luo2025efficient, monemi2025graph} also adopt graph-based learning approaches; however, they overlook the global image-level information inherent to path-planning problems, leaving substantial room for improvement. Regarding hybrid learning–search methods, \cite{joshi2019efficient} employs supervised learning to solve path-planning problems, but it relies on high-quality datasets that are extremely difficult to obtain for path-planning tasks, which limits its scalability. \cite{nair2020solving} integrates neural networks with exact algorithms; although it can achieve high-quality solutions, it still requires substantial computational time when tackling large-scale instances. In addition, approaches such as \cite{ye2023deepaco, ma2021learning, wu2021learning} combine heuristic or meta-heuristic algorithms with reinforcement learning, thereby improving solution quality, yet they still incur noticeable solving times.



To overcome these limitations, we introduce a \emph{\underline{M}ulti\underline{m}odal \underline{F}used \underline{L}earning} (MMFL) framework for solving GTSP in robotic task planning applications. MMFL represents GTSP instances using both a graph structure and a constructed image that encodes the spatial layout of nodes and clusters. By combining these complementary representations, MMFL can better understand the geometric relationships between potential visitation points, leading to improved solution quality.
Our MMFL framework offers a novel perspective on solving GTSP. Unlike conventional approaches that rely solely on graph-based or heuristic methods, MMFL leverages the complementary strengths of both topological and spatial representations. By constructing a rich dual representation that captures both connectivity relationships and geometric distributions of nodes, our model develops a more comprehensive understanding of the problem structure. The framework's architecture dynamically adapts to varying instance scales and spatial configurations, while its fusion mechanism intelligently integrates information streams from different modalities, enabling more context-aware decision making throughout the task planning process.

Through extensive experiments in simulated robotics environments, we demonstrate that our approach outperforms existing methods for GTSP, achieving shorter paths while maintaining computational efficiency suitable for real-time robot operation. Physical robot experiments conducted on mobile vehicle platforms validate that these improvements translate directly to real-world applications. The system's ability to efficiently process environmental information and produce high-quality paths under strict computational constraints makes MMFL a valuable tool for practical robotic applications in warehouses, healthcare facilities, and manufacturing environments.

\section{Preliminary} \label{sec:preliminary}

\noindent\textbf{Problem formulation.}
GTSP can be formally defined on a complete graph $G = (V, E)$, where $V = \left\{v_1, \ldots, v_n\right\}$ represents the set of nodes, and $E = \left\{e(v_i, v_j) \mid v_i, v_j \in V, i \neq j\right\}$ represents the set of edges. The node set $V$ is partitioned into $m$ mutually exclusive and collectively exhaustive clusters $V_1,\ V_2,\ \ldots,V_m$, such that $V=\bigcup_{i=1}^{m}V_i\ and\ V_i\bigcap V_j=\emptyset,\forall i\neq j$.
Each edge $e\left(v_i,v_j\right)\in E$ is associated with a non-negative cost or distance $c_{ij}$. The objective of GTSP is to find a minimum-cost tour that visits exactly one node from each cluster. Formally, we seek a permutation $\pi=\pi_1,\pi_2,\ldots,\pi_m$ of $m$ nodes, where $\pi_i\in V_i$, such that the total tour cost is minimized: $\min_\pi f\left(\pi\right)=\sum_{i=1}^{m-1}c_{\pi_i,\pi_{i+1}}+c_{\pi_m,\pi_1}$. The complete mathematical formulation of the GTSP is provided in Appendix A.




\noindent\textbf{Markov decision process(MDP).} 
Given an GTSP instance $\lambda$, we construct a solution as an MDP. An \emph{agent} iteratively takes the current \emph{state} as input (e.g., the instance information and the partially constructed solution, initially empty), and outputs a probability distribution over candidate nodes belonging to unvisited clusters. The \emph{action} is a node that is greedily selected or randomly sampled from the policy. The \emph{transition} dynamics is joining the node to the partial solution and masking all nodes from its associated cluster to prevent revisits. We parameterize the \emph{policy} $p$ of the agent by a neural network $p_\theta$, so that the probability of constructing a complete tour $\pi$ to the GTSP is expressed by $p_{\theta}(\pi|\lambda) = \prod_{t=1}^{T} p_{\theta} (\pi_{t}|\pi_{<t},\lambda)$, where $\pi_{t}$ and $\pi_{<t}$ represent the selected node and partial solution at the $t$-th step. Typically, the \emph{reward} is defined as the negative value of the total tour cost, e.g., $\mathcal{R}(\pi)= -f(\pi)$.

\section{Multimodal Fused Learning (MMFL) for GTSP} \label{sec:method}
\subsection{Overview}
\begin{figure}[htbp]
    \vspace{-0.3cm}
    \centering 
    \setlength{\abovecaptionskip}{0.cm}
    \includegraphics[width=\textwidth]{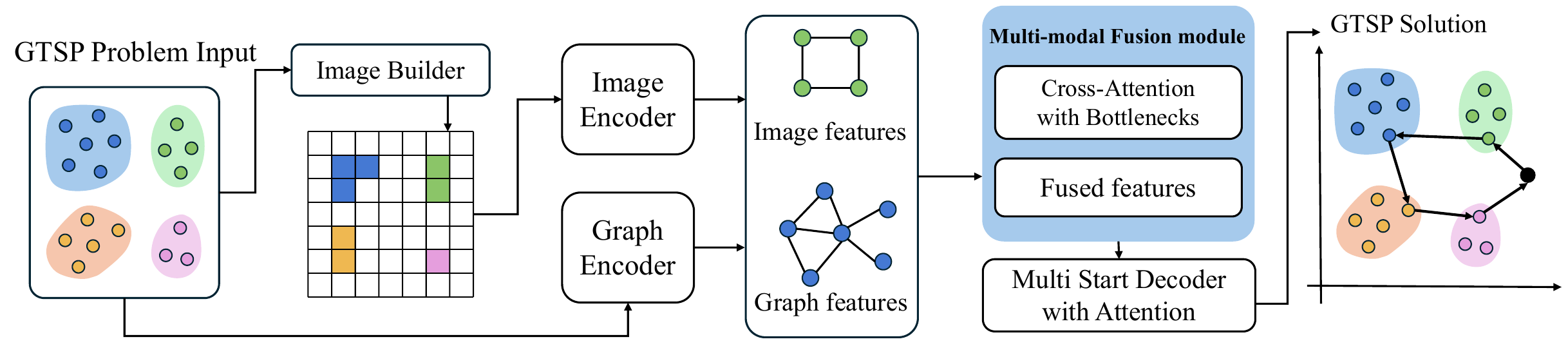} 
    \caption{The overall architecture of MMFL.} 
    \label{architecture}
    \vspace{-0.5cm}
\end{figure}

Our framework incorporates complementary information from both graph and image modalities. The two key challenges are (1) constructing informative images from GTSP instances and (2) effectively fusing graph and image data. To tackle these challenges, we first propose an image construction method with an Adaptive Resolution Scaling (ARS) strategy, and then design specialized graph and image encoders alongside a multimodal fusion module to integrate information from both representations.
As shown in Figure~\ref{architecture}, the overall architecture of MMFL consists of a coordinate-based image builder that transforms GTSP instances into informative image representations, a graph encoder, an image encoder, a multimodal fusion module, and a multi-start decoder. We elaborate on the key components below.

\subsection{Policy Network}

\noindent\textbf{Image builder.}
We construct an informative image representation by encoding both node positions and cluster memberships into a single-channel image. Each GTSP instance consists of $n$ nodes distributed across $m$ clusters, where each node has $2D$ coordinates and a cluster assignment.
Given a node $i$ with coordinates $\left(v_{i,1},v_{i,2}\right) \in \left[ 0,1 \right]^2$ and cluster assignment $c_i \in {1,\ldots,m}$, we map it to a pixel in the image representation as follows. First, we determine the image dimensions $W \times H$ based on the problem size $n$. We then convert the normalized node coordinates to discrete pixel positions: $v_{i,1}^\prime=\left\lfloor v_{i,1}\times W\right\rfloor$ and $v_{i,2}^\prime=\left\lfloor v_{i,2}\times H\right\rfloor$. 
The instance image $I$ is constructed by treating each coordinate as a pixel location and the cluster index as the pixel value:
\begin{align}
    I(x, y) =
        \begin{cases}
        c_i + 1, & \text{if } (x, y) = (v'_{i,1}, v'_{i,2}) \\
        0, & \text{otherwise.}
\end{cases}
\end{align}

When constructing image representations for GTSP instances, a significant challenge arises from varying problem sizes. Using a fixed image resolution would create an inconsistent information density as the number of nodes increases, potentially obscuring important spatial relationships that are crucial to effectively solving the problem.
To address this challenge, we design an ARS strategy, which dynamically adjusts the image resolution based on the problem size. Our ARS strategy uses the formula $W = H = \left\lceil\frac{\alpha\sqrt n}{w}\right\rceil\ \times\ w$, where $n$ is the number of nodes, $w$ is the patch size used in the image encoder, and $\alpha$ is a scaling factor. This formulation ensures that the image dimensions scale proportionally with the square root of the problem size, maintaining an approximately constant node density across different problem scales.

With our ARS strategy, larger problem instances are represented by proportionally larger images, preserving spatial relationships between nodes. For these variable-sized representations, we implement a flexible positional encoding using an MLP that maps coordinates to embeddings:
\begin{align}
    PosEncoder\left(i,j\right)= \text{MLP}\left(\left[i/W,j/H\right]\right),
\end{align}
where $\left(i,j\right)$ are the original patch coordinates in the image grid, and $(i/W,j/H)$ normalizes these coordinates to $\left[0,1\right]^2$.

\noindent\textbf{Image encoder.}
To process the constructed GTSP instance images $I$, we implement a specialized Vision Transformer (ViT) architecture that effectively extracts spatial features while adapting to varying image resolutions. Our image encoder consists of three primary components: patch embedding, positional encoding, and a transformer encoder stack.

The patch embedding layer transforms the instance image into a sequence of patch embeddings. Given an input image $I \in \mathbb{R}^{1 \times H \times W}$, we divide it into non-overlapping patches of size $w \times w$, and project each patch into a $d$-dimensional embedding space using a convolutional operation.

After embedding the patches, we add the position encodings described in the previous section to inject spatial information. The combined embeddings then serve as input to a stack of transformer encoder layers. Each layer consists of a multi-head self-attention (MHSA) sublayer followed by a feed-forward network (FFN), with layer normalization and residual connections:
\begin{align}
    & \widehat{z^{(l)}} = \text{LN}\left( z^{(l-1)} + \text{MHSA}\left( z^{(l-1)} \right) \right) \\
    & z^{(l)} = \text{LN}\left( \widehat{z^{(l)}} + \text{FFN}\left( \widehat{z^{(l)}} \right) \right),
\end{align}
where $z^{(l)}$ represents the patch embeddings at layer $l$, and $\text{LN}$ denotes layer normalization.
The final output of the image encoder is a sequence of enhanced patch embeddings $z^{(L)} \in \mathbb{R}^{N \times d}$, where $N$ is the number of patches and $L$ is the number of transformer layers.

\noindent\textbf{Graph encoder.}
The graph encoder processes the node coordinates and cluster information to generate node embeddings. For each node $i$ in the GTSP instance, we first create an initial embedding by concatenating its coordinates information and cluster index: $u_i=\left[v_{i,1},v_{i,2},c_i\right]$ where $v_{i,1}$ and $v_{i,2}$ represent the node's coordinates, and $c_i$ is the cluster index to which node $i$ belongs.
These features are transformed into initial embeddings through linear projection:
\begin{align}
    h_i^{\left(0\right)}=W_uu_i+b_u,\forall i\in1,\ldots,n,
\end{align}
where $h_i^{\left(0\right)}$ is the initial embedding for node $i$; $W_u$ is a learnable weight matrix; $b_u$ is a bias vector and $n$ is the total number of nodes in the instance.
The graph encoder consists of multiple self-attention layers. Each layer applies MHSA to capture relationships between nodes, followed by an FFN with residual connections and normalization:
\begin{align}
    & h_i^{\left(l\right)}=\mathrm{LayerNorm}\left(h_i^{\left(l-1\right)}+\mathrm{MHSA} \left(h_i^{\left(l-1\right)}\right)\right) \\
    & h_i^{\left(l\right)}=\mathrm{LayerNorm}\left(h_i^{\left(l\right)}+\mathrm{FFN} \left(h_i^{\left(l\right)}\right)\right),
\end{align}
where $h_i^{\left(l\right)}$ is the embedding of node $i$ at layer $l$. This structure enables the model to capture complex relationships between nodes and their cluster assignments.

\noindent\textbf{Multimodal fusion module.}
To effectively integrate information from both graph and image modalities, we implement a bidirectional cross-modal fusion mechanism using learnable bottleneck tokens. This approach enables flexible information exchange while maintaining computational efficiency.

Our fusion mechanism consists of multiple layers that iteratively refine the representations from both modalities. 
Each fusion layer is structured as follows:
\begin{align}
    & \mathbf{G}^{in} = [\mathbf{h}_{graph}; \mathbf{b}_{graph}] \\
    & \mathbf{I}^{in} = [\mathbf{h}_{image}; \mathbf{b}_{image}],
\end{align}
where $\mathbf{h}_{graph} \in \mathbb{R}^{b \times n \times d}$ represents the node embeddings from the graph encoder; $\mathbf{h}_{image} \in \mathbb{R}^{b \times n_m \times d}$ represents the patch embeddings from the image encoder; $\mathbf{b}_{graph} \in \mathbb{R}^{b \times n_b \times d}$ and $\mathbf{b}_{image} \in \mathbb{R}^{b \times n_b \times d}$ are learnable bottleneck parameters specific to each modality. These bottleneck tokens act as information conduits between modalities, with $n_b$ being a hyperparameter controlling fusion capacity.
We then apply the cross-attention to process $\mathbf{G}^{in}$ and $\mathbf{I}^{in}$:
\begin{align}
    & \mathbf{G}^{out} = \text{MHA}(\mathbf{G}^{in}, \mathbf{I}^{in}, \mathbf{I}^{in}) \\
    & \mathbf{I}^{out} = \text{MHA}(\mathbf{I}^{in}, \mathbf{G}^{in}, \mathbf{G}^{in}),
\end{align}
where $\text{MHA}$ denotes the multi-head attention operation. Following the attention operation, we apply layer normalization and feed-forward networks with residual connections:
\begin{align}
    & \mathbf{h}_{graph}^{norm} = \text{LayerNorm}(\mathbf{h}_{graph} + \mathbf{G}^{out}_{:n}) \\
    & \mathbf{b}_{graph}^{norm} = \text{LayerNorm}(\mathbf{b}_{graph} + \mathbf{G}^{out}_{n:}) \\
    & \mathbf{h}_{graph}^{out} = \mathbf{h}_{graph}^{norm} + \text{FFN}(\mathbf{h}_{graph}^{norm}) \\
    & \mathbf{b}_{graph}^{out} = \mathbf{b}_{graph}^{norm} + \text{FFN}(\mathbf{b}_{graph}^{norm}).
\end{align}
Similar operations are also applied to the image features.
After multiple fusion layers, the final fused representation for GTSP is computed as:
\begin{align}
    \mathbf{h}_{fused} = \mathbf{h}_{graph}^{out} + \alpha \cdot \text{Mean}(\mathbf{h}_{image}^{out}),
\end{align}
where $\alpha$ is a weighting factor set to 0.5 in our implementation. This fusion mechanism enables bidirectional information flow between modalities, allowing the model to leverage both the topological information from the graph and the spatial information from the image, resulting in more effective node selection decisions for GTSP.

\noindent\textbf{Multi-start decoder.}
Our multi-start decoder constructs a tour by selecting one node from each cluster in an auto-regressive manner. For each decoding step $t$, we compute the agent's policy based on the embedding of the previously selected node and the fused node representations. At the initial step ($t = 0$), the model always selects the depot node (node 0), as the agent must begin there. At $t = 1$, it selects the $k$-nearest neighbors to the depot, enabling diverse exploration via parallel rollouts~\cite{kwon2020pomo}.
For subsequent steps ($t \geq 2$), we first retrieve the embedding of the previously selected node:
\begin{align}
    h_{\pi_{t-1}} = [h_{\text{fused}}, \pi_{t-1}].
\end{align}
We then compute query vectors by combining node information with global context:
\begin{align}
    q_{\text{graph}} = W_{q_{\text{graph}}}(g), \quad q_{\text{last}} = W_{q_{\text{last}}}(h_{\pi_{t-1}}), \quad q = q_{\text{last}} + q_{\text{graph}},
\end{align}

where $g = h_{\text{graph}} + 0.3 \cdot \text{Mean}(h_{\text{image}})$.
Next, we compute compatibility scores for each node as follows:
\begin{align}
    \alpha_i = \begin{cases}
        -\infty & \text{if node }i\text{ is masked} \\
        C \cdot \tanh\left(\frac{q_c^T W_K h_i}{\sqrt{d}}\right) & \text{otherwise},
        \end{cases}
\end{align}
where $\alpha_i$ is the compatibility score for node $i$; $h_i$ is the fused embedding of node $i$; $W_K$ is a learnable projection matrix; $d$ is the embedding dimension and $C = 10$ is a clipping parameter. A node is masked if it has already been visited or if its cluster has already been visited, enforcing the GTSP constraint.
The policy activation for each node is finally computed using a softmax function:
\begin{align}
    p\left(\pi_t=i\middle|\pi_{1:t-1}\right)=\frac{\exp{\left(\alpha_i\right)}}{\sum_{j}\exp{\left(\alpha_j\right)}},
\end{align}
where $p\left(\pi_t=i\middle|\pi_{1:t-1}\right)$ represents the probability of selecting node $i$ at step $t$ given the previously selected nodes $\pi_{1:t-1}$. During training, nodes are sampled from this policy, while during inference, we can either sample or select greedily based on the configured evaluation mode.

\subsection{Training Procedure}
We train our model using the REINFORCE algorithm~\cite{zhang2021sample}, incorporating a shared baseline~\cite{kwon2020pomo} to reduce variance, which is the average reward across all rollouts for each instance. Specifically, given a batch of $N$ instance $\left\{\lambda_{i}\right\}_{i=1}^{N}$, we sample $k$ tours (i.e., solutions) $\{\pi^{i,j}\}_{j=1}^{k}$ for each instance $\lambda_i$. The model parameters $\theta$ are then updated as: $\nabla_{\theta} \mathcal{L}(\theta|\lambda_i) \simeq \frac{1}{kN}{\sum}_{i=1}^{N}{\sum}_{j=1}^{k}(\mathcal{R}(\pi^{i,j}|\lambda_i) -b(\lambda_{i}))\triangledown_{\theta}\log p_{\theta}(\pi^{i,j}|\lambda_{i})$, where $b(\lambda_i)$ denotes the shared baseline for instance $\lambda_i$, computed as: $b(\lambda_i)=\frac{1}{k} \sum_{j=1}^{k} \mathcal{R} (\pi^{i,j})$.
Detailed information on the training procedure, including the algorithm and dataset, is provided in Appendix B. 


\section{Experiments} \label{sec:experiments}

\subsection{Experimental Settings}
\noindent\textbf{Problems.} We conduct extensive experiments to evaluate the performance of our proposed MMFL framework on GTSP. For comprehensive evaluation, we generate five scale-based GTSP instance sets: $(n = 20, m = 4), (n = 50, m = 10), (n = 100, m = 20), (n = 150, m = 30), (n = 200, m = 40)$, where $n$ represents the number of nodes and $m$ the number of groups. In these instances, node coordinates were uniformly sampled from the unit square $[0,1]^2$. To assess algorithmic generalization, we create four distinct grouping configurations using $n = 100$ nodes: Random Groups (random assignment), Proximity-Based Groups (nearest centroid assignment), Density-Based Groups (DBSCAN-style assignment), and Hybrid Groups (mixed assignment strategies). Additionally, we examine four group size distributions with $n = 100$ total nodes: Uniform Groups (20 groups, 5 nodes each), Small Groups (40 groups, 2-3 nodes each), Large Groups (10-12 groups, 8-10 nodes each), and Mixed Groups (15-20 groups, 1-15 nodes each). For each setting, we generate 30 GTSP instances to form the test dataset.

\noindent\textbf{Hyperparameters.} For model training and evaluation, we design our neural network with an embedding dimension of 128, 3 image and graph encoder layers, and 8 attention heads. The image processing branch uses 16×16 patch size with ARS strategy. Our multimodal fusion module consists of 3 fusion layers with 10 bottleneck tokens per modality. We define the number of rollouts sampled for each instance as one quarter of the problem size (i.e., \( k = \text{int}(n/4) \)). The model parameters are optimized using Adam with a learning rate of $1 \times 10^{-4}$ and a weight decay of $1 \times 10^{-6}$. A cosine annealing scheduler is employed to gradually decrease the learning rate during training. The model is trained for 200 epochs with gradient clipping set to 1.0. Each epoch processes 100,000 instances with a batch size of 128.

\noindent\textbf{Baselines.} We compare our approach against several state-of-the-art GTSP algorithms: Lin-Kernighan-Helsgaun (LKH), a heuristic typically producing near-optimal solutions for TSP variants~\cite{helsgaun2015solving}; Google OR-Tools, an efficient optimization toolkit with specialized routing capabilities~\cite{cuvelier2023or}; MA, combining genetic algorithms with local search techniques~\cite{cosma2024novel}; Adaptive Large Neighborhood Search (ALNS), a destroy-and-repair metaheuristic~\cite{smith2017glns}; and Policy Optimization with Multiple Optima (POMO), a reinforcement learning method exploiting problem symmetry~\cite{kwon2020pomo} for vehicle routing problems. All methods are executed on a machine equipped with an RTX 4080 Laptop GPU and an Intel i9-13980HX CPU.

\noindent\textbf{Metrics.} We evaluate algorithm performance using three metrics: the average objective value (Obj.) representing the total tour length; the average optimality gap (Gap) calculated as (Obj - Best)/Best × 100\%; and total run time (Time) per test dataset measured in CPU seconds. For learning-based methods, we only report inference time. The best results are highlighted in bold.

\begin{table}[t]
  \centering
  \caption{Comparison of algorithms on GTSP with various problem sizes.}
  \setlength{\tabcolsep}{4pt}
  \renewcommand{\arraystretch}{1.05}
  \resizebox{\linewidth}{!}{
  \begin{tabular}{c *{5}{ccc}}
    \toprule
    \multirow{2}{*}{Algorithm}
      & \multicolumn{3}{c}{$n{=}20,\,m{=}4$}
      & \multicolumn{3}{c}{$n{=}50,\,m{=}10$}
      & \multicolumn{3}{c}{$n{=}100,\,m{=}20$}
      & \multicolumn{3}{c}{$n{=}150,\,m{=}30$}
      & \multicolumn{3}{c}{$n{=}200,\,m{=}40$} \\ 
    \cmidrule(lr){2-4}\cmidrule(lr){5-7}\cmidrule(lr){8-10}
    \cmidrule(lr){11-13}\cmidrule(lr){14-16}
    & Obj. & Gap & Time
    & Obj. & Gap & Time
    & Obj. & Gap & Time
    & Obj. & Gap & Time
    & Obj. & Gap & Time \\
    \midrule
    LKH      & 1.18 & 4.42 & 0.60  & 1.98 & 15.79 & 2.10 & 2.73 & 21.33 & 6.90 & 3.33 & 4.06 & 30.30 & 3.83 & 5.51 & 42.00 \\
    OR-Tools & 1.24  & 9.73  & 9.35  & 2.13  & 24.56  & 13.08  & 2.84  & 26.22  & 18.21  & 3.69  & 15.31  & 27.54  & 3.78  & 4.13  & 45.68  \\
    MA       & \textbf{1.13} & \textbf{0.00} & 1.80 & 1.92 & 12.28 & 9.30 & 2.71 & 20.44 & 106.80 & 3.40 & 6.25 & 655.20 & 3.85 & 6.06 & 1791.30 \\
    ALNS     & 1.18 & 4.42 & 1.20  & 1.94 & 13.45 & 2.10 & 2.64 & 17.33 & 6.30 & 3.23 & 0.94 & 69.60 & 3.69 & 1.65 & 33.00 \\
    POMO     & \textbf{1.13} & \textbf{0.00} & 0.11 & \textbf{1.71} & \textbf{0.00} & 0.12 & 2.72 & 20.89 & 0.12 & 3.32 & 3.75 & 0.13 & 3.77 & 3.86 & 0.13 \\
    MMFL     & \textbf{1.13} & \textbf{0.00} & 0.12 & \textbf{1.71} & \textbf{0.00} & 0.11 & \textbf{2.25} & \textbf{0.00} & 0.13 & \textbf{3.20} & \textbf{0.00} & 0.13 & \textbf{3.63} & \textbf{0.00} & 0.14 \\
    \bottomrule
  \end{tabular}}
  \label{table:gtsp-comparison}
  \vspace{-4mm}
\end{table}

\subsection{Experiment Results}
\noindent\textbf{Comparison analysis.} 
Our comparison results are shown in Table~\ref{table:gtsp-comparison}, the results demonstrate the superior performance of our MMFL framework across all problem scales. For small instances, MMFL matches the performance of MA and POMO, but as problem size increases, MMFL's advantage becomes substantial, improving over LKH by 4.06 - 21.33\% and OR-Tools by 4.13 - 26.22\%. MMFL also maintains efficient inference times, comparable to POMO but significantly faster than MA and LKH whose computational costs increase dramatically with problem size.

\begin{table}[ht]
  \vspace{-3mm}
  \centering
  \caption{Performance on different GTSP instance types ($n{=}100$).}
  \setlength{\tabcolsep}{4pt}
  \renewcommand{\arraystretch}{1.05}
  \resizebox{0.82\linewidth}{!}{
  \begin{tabular}{c *{4}{ccc}}
    \toprule
    \multirow{2}{*}{Algorithm}
      & \multicolumn{3}{c}{Random}
      & \multicolumn{3}{c}{Proximity}
      & \multicolumn{3}{c}{Density}
      & \multicolumn{3}{c}{Hybrid} \\ 
    \cmidrule(lr){2-4}\cmidrule(lr){5-7}\cmidrule(lr){8-10}\cmidrule(lr){11-13}
      & Obj. & Gap & Time
      & Obj. & Gap & Time
      & Obj. & Gap & Time
      & Obj. & Gap & Time \\
    \midrule
    LKH      & 3.04 & 28.27 & 6.90  & 3.21 & 1.90  & 6.70 & \textbf{1.45} & \textbf{0.00} & 9.00 & 2.53 & 17.13 & 9.40 \\
    OR-Tools & 3.24  & 36.71  & 40.23  & 3.92  & 24.44  & 32.98  & 1.80  & 22.45  & 20.29  & 2.77  & 28.24  & 30.27  \\
    MA       & 3.29 & 38.82 & 84.30 & 3.24 & 2.86  & 62.70& 1.46 & 0.69  & 61.50& 2.58 & 19.44 & 71.10 \\
    ALNS     & 2.97 & 25.32 & 6.90  & 3.22 & 2.22  & 5.70 & \textbf{1.45} & \textbf{0.00} & 4.50 & 2.53 & 17.13 & 5.70 \\
    POMO     & 2.46 & 3.80  & 0.12  & 3.20 & 1.59  & 0.14 & 1.56 & 7.59  & 0.13 & 2.42 & 12.04 & 0.14 \\
    MMFL     & \textbf{2.37} & \textbf{0.00} & 0.13 & \textbf{3.15} & \textbf{0.00} & 0.15 & 1.47 & 1.38 & 0.09 & \textbf{2.16} & \textbf{0.00} & 0.12 \\
    \bottomrule
  \end{tabular}}
  \label{table:instance-types}
\end{table}
\FloatBarrier 
\begin{table}[htbp]
 \vspace{-5mm}
  \centering
  \caption{Performance on GTSP with varying group sizes ($n{=}100$).}
  \setlength{\tabcolsep}{4pt}
  \renewcommand{\arraystretch}{1.05}
  \resizebox{0.82\linewidth}{!}{
  \begin{tabular}{c *{4}{ccc}}
    \toprule
    \multirow{2}{*}{Algorithm}
      & \multicolumn{3}{c}{Uniform}
      & \multicolumn{3}{c}{Small}
      & \multicolumn{3}{c}{Large}
      & \multicolumn{3}{c}{Mixed} \\ 
    \cmidrule(lr){2-4}\cmidrule(lr){5-7}\cmidrule(lr){8-10}\cmidrule(lr){11-13}
      & Obj. & Gap & Time
      & Obj. & Gap & Time
      & Obj. & Gap & Time
      & Obj. & Gap & Time \\
    \midrule
    LKH      & 2.24 & 6.67  & 6.60 & 4.23 & 7.91  & 21.90 & 1.74 & 45.00 & 2.70 & 2.29 & 13.93 & 4.50 \\
    OR-Tools & 2.62 & 24.76 & 38.56 & 4.78 & 21.94 & 62.35 & 1.81 & 50.83 & 20.56 & 2.84 & 41.29 & 40.37 \\
    MA       & 2.47 & 17.62 & 78.30 & 4.38 & 11.73 & 1026.00 & 1.65 & 37.50 & 15.90 & 2.29 & 13.93 & 44.70 \\
    ALNS     & 2.24 & 6.67  & 6.90 & 4.21 & 7.40  & 30.62 & 1.74 & 45.00 & 3.10 & 2.22 & 10.45 & 4.51 \\
    POMO     & 2.21 & 5.24  & 0.12 & 4.12 & 5.10  & 0.14 & 1.63 & 35.83 & 0.13 & 2.36 & 17.41 & 0.14 \\
    MMFL     & \textbf{2.10} & \textbf{0.00} & 0.12 & \textbf{3.92} & \textbf{0.00} & 0.16 & \textbf{1.20} & \textbf{0.00} & 0.12 & \textbf{2.01} & \textbf{0.00} & 0.13 \\
    \bottomrule
  \end{tabular}}
  \label{table:group-size}
  \vspace{-3mm}
\end{table}

\begin{figure}[htbp]
  \centering
  \begin{subfigure}[b]{\textwidth}
    \centering
    \includegraphics[width=\textwidth]{table3.pdf}
    \caption{GTSP instances with different grouping distributions.}
    \label{fig:grouping}
  \end{subfigure}
  \vspace{0.4cm}
  \begin{subfigure}[b]{\textwidth}
    \centering
    \includegraphics[width=\textwidth]{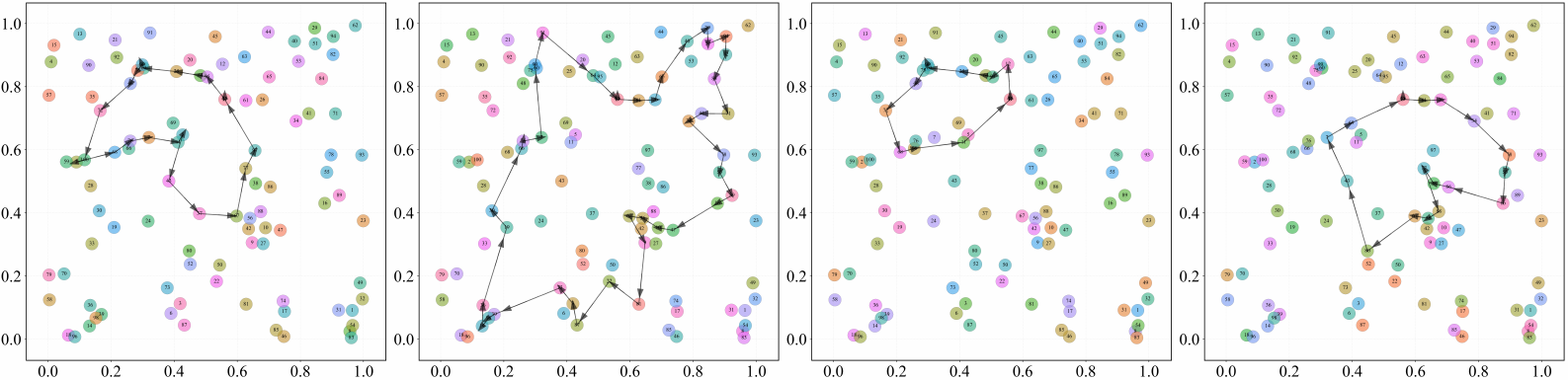}
    \caption{GTSP instances with varying group sizes.}
    \label{fig:groupsize}
  \end{subfigure}
  \vspace{-0.8cm}
  \caption{GTSP instance visualizations under different configurations.}
  \label{fig:gtsp-configs}
\end{figure}

\noindent\textbf{Generalization analysis.}
Tables~\ref{table:instance-types} and \ref{table:group-size} demonstrate MMFL's generalization across different problem structures. MMFL achieves best performance in 3 of 4 group distributions, with only a slight 1.38\% gap in Density-Based Groups where LKH and ALNS perform best. For varying group sizes, MMFL also consistently delivers superior solutions, with substantial improvements in Large Groups which is 35.83\% better than POMO and 45.00\% better than LKH. In Small Groups, MMFL is also 5.10\% better than POMO, 7.40\% better than ALNS. 
Figure~\ref{fig:grouping} and Figure~\ref{fig:groupsize} show route maps for some examples, these results confirm that MMFL effectively generalizes across diverse problem structures with different spatial distributions and group configurations, while maintaining computational efficiency.


 \begin{wrapfigure}{r}{0.4\textwidth}
 \centering
    \vspace{-0.5cm}
    \setlength{\abovecaptionskip}{0.cm}
  \includegraphics[width=0.4\textwidth]{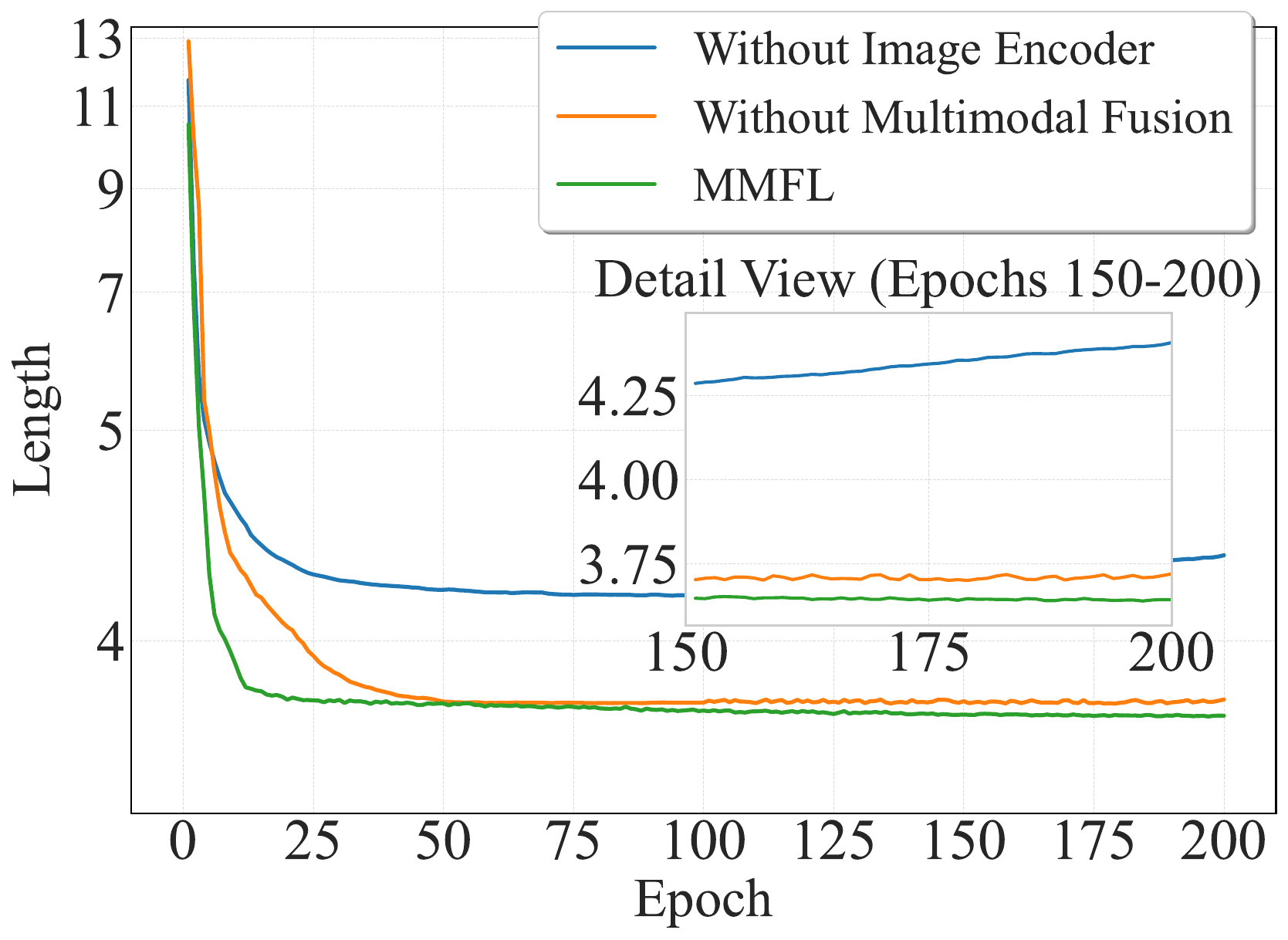}
  \caption{Convergence curves.}
  \label{convergence}
  \vspace{-0.5cm}
\end{wrapfigure}

\noindent\textbf{Ablation study.} To validate each component's contribution, we conducted ablation studies on the image encoder and multimodal fusion module. Figure~\ref{convergence} shows our complete MMFL model consistently maintains lower length values than ablated versions. Without our image encoder, convergence is slower in early epochs, while removing multimodal fusion causes higher initial values and instability. Even after convergence, MMFL still retains a consistent advantage of approximately 0.1 in tour length, confirming both components are essential for optimal performance. Detailed results are provided in Appendix C.

\begin{figure}[htbp]
  \centering
  \begin{minipage}[c]{0.38\textwidth}
    \centering
    \includegraphics[height=4cm,trim=3mm 20mm 3mm 10mm,clip]{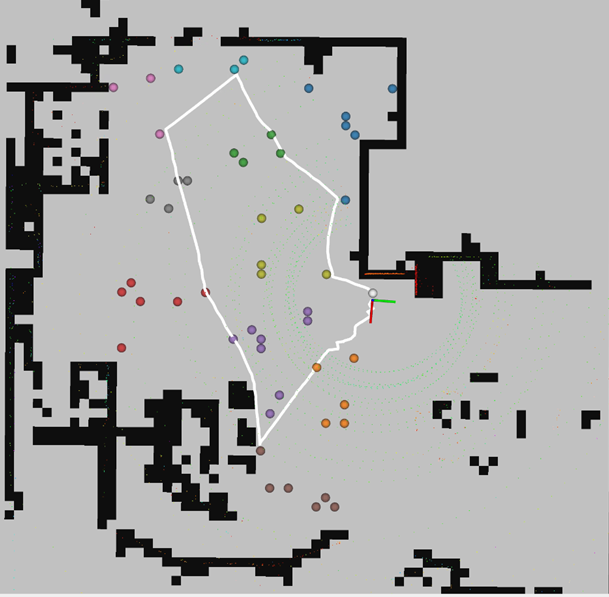}
    \caption{MMFL solution path for a multi-zone exploration task.}
    \label{Fig2}
  \end{minipage}
  \hfill
  \begin{minipage}[c]{0.60\textwidth}
    \centering
    \vspace{-4mm}
      \includegraphics[height=4cm, trim=3mm 20mm 3mm 10mm, clip]{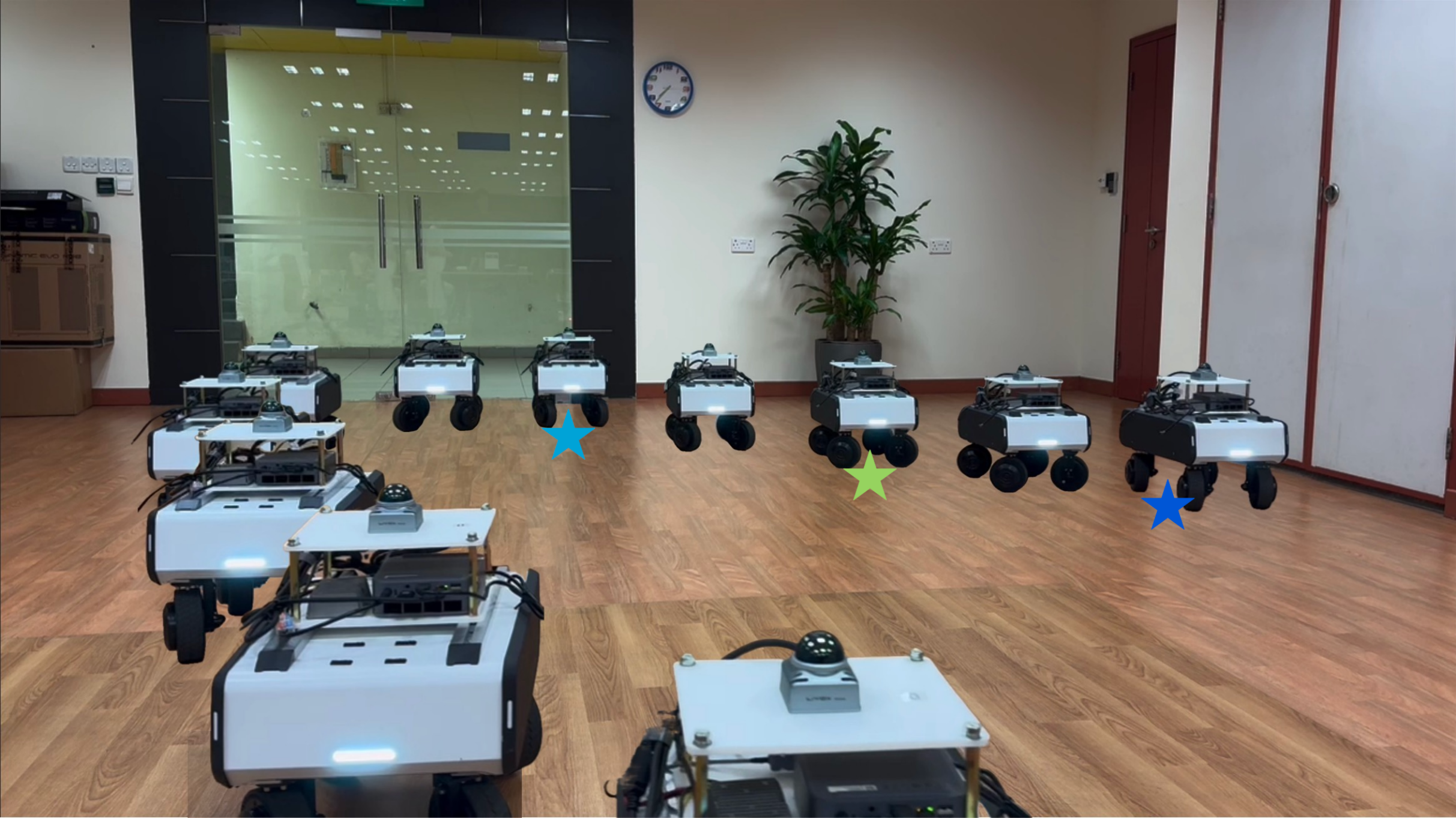}
    \caption{Locomotion in a real-world environment.}
    \label{fig:physical}
  \end{minipage}
  \vspace{-4mm}
\end{figure}

\noindent\textbf{Experimental Validation.}
We implemented our proposed MMFL framework on a physical robot platform to validate its effectiveness in real-world task planning scenarios. We deployed our algorithm on a mobile robot operating in a complex indoor environment where the robot needed to explore different regions, each containing multiple possible observation points. Figures~\ref{Fig2} and ~\ref{fig:physical} show our experimental results, where the robot successfully navigated through multiple zones (indicated by different colored nodes). The white path represents the solution generated by our algorithm, demonstrating MMFL's ability to effectively solve GTSP instances in practical settings.



\section{Conclusion} \label{sec:conclu}
In this paper, we introduce MMFL, a novel framework for solving GTSP in robotic task planning scenarios. Our method uniquely integrates both graph-based and spatial representation learning through a carefully designed fusion architecture incorporating ARS strategy and dedicated bottleneck mechanisms. Extensive experimental results across multiple problem sizes and configurations demonstrate that MMFL consistently outperforms state-of-the-art algorithms, achieving optimal or near-optimal solutions with significantly shorter tour lengths while maintaining competitive computational efficiency. Our performance advantages are particularly pronounced for complex instances with diverse spatial distributions and group configurations, highlighting MMFL's strong generalization capabilities. Physical robot experiments further validate the practical effectiveness of our approach in real-world navigation scenarios. While some challenges remain for extreme problem distributions, we believe that MMFL represents a significant advancement in solving GTSP efficiently for robotic applications, with potential extensions to other combinatorial optimization problems involving spatial reasoning.

\clearpage

\section{Limitation} \label{sec:limits}
While our MMFL framework demonstrates strong performance in solving GTSP instances for robotic task planning, several limitations should be acknowledged:

\begin{enumerate}[label={(\arabic*)}]
    \item \textbf{Limited generalization to novel distributions.} The current model is trained on a fixed distribution of problem instances and may exhibit degraded performance when faced with significantly different distributions, such as asymmetric node layouts, highly imbalanced cluster densities, or non-uniform spatial patterns. In the future, we will boost the robustness of MMFL via robust learning technologies (e.g., meta-learning, curriculum learning, and few-shot fine-tuning).
    \item \textbf{Computational scalability challenges.} Although MMFL incorporates our ARS strategy to improve efficiency, solving very large-scale GTSP instances remains computationally expensive. Our multimodal fusion module, while effective, introduces additional overhead compared to graph-only baselines, particularly when scaling to instances with thousands of nodes and clusters. 
    Future work will explore integrating a divide-and-conquer architecture to maintain both high solution quality and computational efficiency in large-scale task planning scenarios.

    \item \textbf{Environmental complexity limitations.} Our current framework assumes a static environment with known node positions. However, in real-world scenarios, robots often operate in dynamic and uncertain environments with moving obstacles, temporal constraints, and localization uncertainty caused by challenging terrain or infrastructure limitations. Future work will focus on incorporating such dynamic elements into the planning process.
\end{enumerate}



\bibliography{example}  

\newpage

\appendix

\vbox{
\hrule height 4pt
\vskip 0.25in
\vskip -\parskip%
\centering
{\LARGE\bf Appendix}
\vskip 0.29in
\vskip -\parskip
\hrule height 1pt
}

\section{Mathematical model of GTSP}
This appendix provides the detailed integer linear programming(ILP) formulation of the Generalized Traveling Salesman Problem (GTSP) referened in the main body of this paper.

\noindent\textbf{Definitions and notation.} The GTSP is defined on a graph where nodes are partitioned into a number of predefined, mutually exclusive clusters. The objective is to find a minimum-cost tour that visits exactly one node from each cluster.

Let $V = {v_1, v_2, \dots, v_n}$ be the set of $n$ nodes. The set of nodes $V$ is partitioned into $m$ mutually exclusive and collectively exhaustive clusters $V_1, V_2, \dots, V_m$, such that $V=\bigcup_{i=1}^{m}V_i\ and\ V_i\bigcap V_j=\emptyset,\forall i\neq j$. Let $c_{ij}$ be the non-negative cost (e.g., distance or travel time) associated with traversing the arc from node $i \in V$ to node $j \in V$.

The following decision variables are used in the model:

$x_{ij}$: A binary variable. $x_{ij} = 1$ if the tour travels directly from node $i$ to node $j$, and 0 otherwise.

$y_i$: A binary variable. $y_i = 1$ if node $i$ is included in the tour, and 0 otherwise.

$u_i$: An auxiliary continuous variable used for Miller-Tucker-Zemlin (MTZ) subtour elimination. If node $i$ is visited, it represents the position of node $i$ in the sequence of the tour.

\noindent\textbf{Objective Function} The objective is to minimize the total cost of the tour, which is the sum of the costs of all selected arcs:
\begin{equation}
\begin{split}
    \min Z = \sum_{i \in V} \sum_{j \in V, i \neq j} c_{ij} x_{ij}
\end{split}
\end{equation}

\noindent\textbf{Constraints} The minimization of the objective function is subject to the following constraints.
\begin{align}
    & \sum_{i \in V_p} y_i = 1, \forall p \in \{1, 2, \dots, m\}  \label{2} \\
    & \sum_{j \in V, j \neq i} x_{ji} = y_i, \forall i \in V  \label{3} \\
    & \sum_{j \in V, j \neq i} x_{ij} = y_i, \forall i \in V  \label{4} \\
    & y_i \le u_i \le m \cdot y_i, \forall i \in V  \label{5} \\
    & u_i - u_j + m \cdot x_{ij} \le m - 1, \forall i, j \in V, i \neq j \label{6} \\ 
    & x_{ij} \in \{0, 1\}, \forall i, j \in V, i \neq j \label{7} \\ 
    & y_i \in \{0, 1\}, \forall i \in V \label{8} \\
    & u_i \geq 0, \forall i \in V \label{9} 
\end{align}

Constraint(\ref{2}) ensures that exactly one node is selected and included in the final tour from each of the $m$ clusters. Constraint(\ref{3}) and (\ref{4}) ensure that if a node $i$ is selected for the tour (i.e., $y_i = 1$), then exactly one arc must enter that node, and exactly one arc must leave it. If a node is not selected ($y_i = 0$), no arcs can be incident to it, this guarantees proper connectivity at the node level. Constraints(\ref{5}) and (\ref{6}) are the MTZ subtour elimination constraints, designed to ensure a single, continuous tour. Constraint(\ref{5}) represents that if node $i$ is visited ($y_i=1$), $u_i$ takes a value between $1$ and $m$ (representing its position in the $m$-node tour), and $u_i=0$ if node $i$ is not visited. Constraint(\ref{6}) then imposes a sequential order: if the tour travels from node $i$ to node $j$ ($x_{ij}=1$), this implies $u_j \ge u_i + 1$, meaning node $j$ must appear after node $i$ in the sequence, thereby preventing the formation of premature cycles or subtours before all $m$ clusters have been visited. Constraints(\ref{7} - \ref{9}) are the variable constraints.

The GTSP is an NP-hard combinatorial optimization problem. This implies that finding a provably optimal solution using exact algorithms based on the above ILP formulation can be computationally intractable for large-scale instances. This intractability underscores the necessity for advanced approaches, such as Deep Reinforcement Learning, to effectively tackle such complex problems by learning high-quality solution policies.

\section{Training Procedure}
The training algorithm for our MMFL framework is summarized in Algorithm~\ref{alg:reinforce}. We employ the REINFORCE algorithm to train our model. To reduce the variance of the policy gradient and stabilize the training process, we incorporate a shared baseline. This baseline is computed for each problem instance by averaging the rewards obtained from multiple rollouts (i.e., multiple generated solution tours). For each GTSP instance $\lambda_i$, we sample $k$ different routes $\pi^{i,j}$ using the \textsc{SampleRollout} function (line 4), which utilizes the multi-start decoder to generate feasible solutions from different starting nodes based on the $k$-nearest neighbors principle. We then compute the shared baseline $b(\lambda_i)$ for each instance, which is the average reward across the $k$ solutions (line 5).

In line 7, we calculate the policy gradient $\nabla_{\theta} \mathcal{L} (\theta | \lambda_i)$ using the core REINFORCE formula. Here, $\mathcal{R} (\pi^{i,j} | \lambda_i) - b(\lambda_i)$ serves as an advantage function, reducing the variance of the gradient estimates. The $\nabla_{\theta} \mathcal{L} (\theta | \lambda_i)$ term indicates increasing the probability of producing solutions with high rewards while decreasing the probability of solutions with low rewards. Finally, line 8 updates the model parameters $\theta$ using the Adam optimizer, adjusting the weights through the policy gradient and the learning rate $\alpha$.

\begin{algorithm}[H]
\SetAlgoLined
\caption{Training Algorithm for MMFL}
\label{alg:reinforce}
\KwIn{Initialize policy network $p_\theta$ with random weights $\theta$, number of training epochs $E$, number of rollouts $k$, number of batches $B$ per epoch, batch size $N$, learning rate $\alpha$}
\For{epoch = 1, \dots, E}{
    \For{b = 1, \dots, B}{
        \For{i = 1, ..., N}{
            $\pi^{i,j} \leftarrow$ \textsc{SampleRollout}($\lambda_i$) \quad $\forall j \in \{1, 2, \dots, k\}$\;
            $b(\lambda_i) \leftarrow \frac{1}{k} \sum_{j=1}^{k} \mathcal{R}(\pi^{i,j})$\;
        }
        $\nabla_{\theta} \mathcal{L}(\theta|\lambda_i) \simeq \frac{1}{kN}{\sum}_{i=1}^{N}{\sum}_{j=1}^{k}(\mathcal{R}(\pi^{i,j}|\lambda_i) -b(\lambda_{i}))\triangledown_{\theta}\log p_{\theta}(\pi^{i,j}|\lambda_{i})$ \;
        $\theta \leftarrow \theta + \alpha\nabla_{\theta} \mathcal{L}(\theta|\lambda_i)$ \;
    }
}
\end{algorithm}

\section{Ablation Study}
Our ablation study demonstrates the critical contribution of each component in the MMFL framework for solving GTSP. The results are shown in Table~\ref{ablation}. For smaller problem instances ($n=20, c=4$ and $n=50, c=10$), all model variants perform similarly, but as problem complexity increases, the advantages of our complete architecture become evident. Without the Image Encoder, performance gaps range from 1.59\% to 10.00\%, while removing the Multimodal Fusion module causes even larger degradation (up to 35.83\% for Large group distributions).

The most significant performance differences appear in complex spatial configurations and heterogeneous structures (Hybrid, Mixed, and Large groups), confirming that MMFL's strength lies in effectively integrating both topological and geometric information. Importantly, all model variants maintain comparable inference times, demonstrating that our approach achieves superior solution quality without sacrificing the computational efficiency required for real-time robotic task planning.
\begin{table}[htbp]
\vspace{-5mm}
  \centering
  \caption{Ablation study.}
  \setlength{\tabcolsep}{4pt}
  \renewcommand{\arraystretch}{1.05}
  \resizebox{0.82\linewidth}{!}{
  \begin{tabular}{c *{9}{p{0.09\linewidth}}}
    \toprule
    \multirow{2}{*}{Instances}
      & \multicolumn{2}{c}{MMFL}
      & \multicolumn{3}{c}{w/o Multimodal Fusion}
      & \multicolumn{3}{c}{w/o Image Encoder} \\
    \cmidrule(lr){2-3}\cmidrule(lr){4-6}\cmidrule(lr){7-9}
      & Obj. & Time
      & Obj. & Gap & Time
      & Obj. & Gap & Time \\
    \midrule
    $n=20, c=4$ &    \textbf{1.13} & 0.12  & 1.13  & 0.00  & 0.11  & 1.13  & 0.00  & 0.12  \\
    $n=50, c=10$ &   \textbf{1.71} & 0.11  & 1.71  & 0.00  & 0.12  & 1.71  & 0.00  & 0.09  \\
    $n=100, c=20$ &  \textbf{2.25} & 0.13  & 2.31  & 2.67  & 0.11  & 2.69  & 19.56  & 0.11  \\
    $n=150, c=30$ &  \textbf{3.20} & 0.13  & 3.32  & 3.75  & 0.13  & 3.53  & 10.31  & 0.12  \\
    $n=200, c=40$ &  \textbf{3.63} & 0.14  & 3.74  & 3.03  & 0.13  & 3.78  & 4.13  & 0.12  \\
    \midrule
    Random        &  \textbf{2.37 } & 0.13  & 2.41  & 1.69  & 0.12  & 2.47  & 4.22  & 0.12  \\
    Proximity     &  \textbf{3.15 } & 0.15  & 3.20  & 1.59  & 0.13  & 3.22  & 2.22  & 0.12  \\
    Density       &  \textbf{1.47 } & 0.09  & 1.50  & 2.04  & 0.10  & 1.56  & 6.12  & 0.08  \\
    Hybrid        &  \textbf{2.16 } & 0.12  & 2.27  & 5.09  & 0.11  & 2.42  & 12.04  & 0.10  \\
    \midrule
    Uniform       &  \textbf{2.10 } & 0.12  & 2.23  & 6.19  & 0.12  & 2.31  & 10.00  & 0.10  \\
    Small         &  \textbf{3.92 } & 0.16  & 3.98  & 1.53  & 0.14  & 4.12  & 5.10  & 0.14  \\
    Large         &  \textbf{1.20 } & 0.12  & 1.32  & 10.00  & 0.11  & 1.63  & 35.83  & 0.10  \\
    Mixed         &  \textbf{2.01 } & 0.13  & 2.21  & 9.95  & 0.13  & 2.36  & 17.41  & 0.11  \\
    \bottomrule
  \end{tabular}}
  \label{ablation}
\end{table}


\end{document}